\documentclass[3p,times]{elsarticle}
\usepackage[hidelinks]{hyperref}

\usepackage{array}
\usepackage{booktabs}
\newcolumntype{E}[1]{>{\raggedright\arraybackslash}p{#1}}
\newcolumntype{R}[1]{>{\centering\arraybackslash}p{#1}}
\newcolumntype{T}[1]{>{\raggedleft\arraybackslash}p{#1}}
\newcolumntype{D}[1]{>{\raggedright\arraybackslash}m{#1}}
\newcolumntype{F}[1]{>{\centering\arraybackslash}m{#1}}
\newcolumntype{G}[1]{>{\raggedleft\arraybackslash}m{#1}}
\newcolumntype{C}[1]{>{\raggedright\arraybackslash}b{#1}}
\newcolumntype{V}[1]{>{\centering\arraybackslash}b{#1}}
\newcolumntype{B}[1]{>{\raggedleft\arraybackslash}b{#1}}

\usepackage{xcolor}
\definecolor{deepcarrotorange}{rgb}{0.91, 0.41, 0.17}
\definecolor{lightgray}{gray}{0.95}
\usepackage{xparse}
\NewDocumentCommand{\codeword}{v}{%
\texttt{\colorbox{lightgray}{#1}}%
}
\journal{Pattern Recognition}

\begin{document}
\begin{frontmatter}

\title{Automatic analysis of artistic paintings using information-based measures}


\author[add1,add2]{Jorge Miguel Silva}\corref{mycorrespondingauthor}
\ead{jorge.miguel.ferreira.silva@ua.pt}

\author[add1,add2,add3]{Diogo Pratas}
\author[add1,add2]{Rui Antunes}
\author[add1,add2]{S\'ergio Matos}
\author[add1,add2]{Armando J. Pinho}

\address[add1]{Institute of Electronics and Informatics Engineering of Aveiro, University of Aveiro, Portugal}
\address[add2]{Department of Electronics, Telecomunications and Informatics, University of Aveiro, Portugal}
\address[add3]{Department of Virology, University of Helsinki, Finland}

\cortext[mycorrespondingauthor]{Corresponding author}

\begin{abstract}
The artistic community is increasingly relying on automatic computational analysis for authentication and classification of artistic paintings.
In this paper, we identify hidden patterns and relationships present in artistic paintings by analysing their complexity, a measure that quantifies the sum of characteristics of an object.
Specifically, we apply Normalized Compression (NC) and the Block Decomposition Method (BDM) to a dataset of 4,266 paintings from 91 authors and examine the potential of these information-based measures as descriptors of artistic paintings. 
Both measures consistently described the equivalent types of paintings, authors, and artistic movements.
Moreover, combining the NC with a measure of the roughness of the paintings creates an efficient stylistic descriptor.
Furthermore, by quantifying the local information of each painting, we define a fingerprint that describes critical information regarding the artists' style, their artistic influences, and shared techniques. More fundamentally, this information describes how each author typically composes and distributes the elements across the canvas and, therefore, how their work is perceived.
Finally, we demonstrate that regional complexity and two-point height difference correlation function are useful auxiliary features that improve current methodologies in style and author classification of artistic paintings.
The whole study is supported by an extensive website (\href{http://panther.web.ua.pt}{http://panther.web.ua.pt}) for fast author characterization and authentication.\end{abstract}


\end{frontmatter}

\section*{Introduction}
Artistic paintings are concrete visual expressions of human evolution and creativity to share emotions, values, visions, beliefs, and trends of history and culture. The creation, interpretation, and analysis of artistic paintings are social, contextual, subjective, passive, and, beyond superficial characteristics, complex to compute and automatize \cite{Weisberg-2006a}. In particular, it is theorized that art is an output of social agents, particularly a human experience, that can only be imitated by machines \cite{Hertzmann-2018a}.

One of the non-trivial characteristic analysis of artistic paintings is related to the process of measuring the information contained in those paintings. Artistic paintings contain information related to schools, periods, and artists \cite{Khan-2014a}. The artistic community widely uses automatic computational analysis of artistic paintings for authentication of artistic paintings \cite{Lyu-2004a,Kim-2014a}. Currently, this process does not substitute human experts completely; however, it is an essential additional control for fraud and mislead detections \cite{Zhang-2017b}. Furthermore, applying new techniques and pre-existing ones that are new to the field, can be useful not only for authorship attribution and fraud detection but also for art style categorization and organization, and even for art content explanation.

In this paper, we introduce novel solutions for automatic computational analysis of artistic paintings and for the problem of artist authentication. When addressing artist authentication, several questions arise: What defines a painter's style? How does the author expose information? How does the author differs and relates to other artists?
Furthermore, taking inspiration from information theory: How do we best quantify information in a painting? How is the information utilized across the canvas? Moreover, what can information quantification tell us about the author's style, way of painting, and relationships with other authors?
These complex questions are at the core of this paper's development, where we describe and compare solutions for unsupervised measures of probabilistic and algorithmic information in images (2D) of artistic paintings.

  Our contributions are as follows:
\begin{itemize}
\item We perform a direct comparison between state-of-the-art unsupervised probabilistic and algorithmic information measures to specify each measure's strengths and weaknesses.
\item We show that hidden patterns and relationships present in artistic paintings can be identified by analysing their complexity.
\item We show an efficient stylistic descriptor by combining the Normalized Compression and a measure of the paintings' roughness.
\item We propose a new descriptor of the artists' style, artistic influences, and shared techniques.
\item We show that average local complexity describes how each author typically composes and distributes the elements across the canvas and, therefore, how their work is perceived.
\item We demonstrate that these measures can serve as useful auxiliary features capable of improving current methodologies in the classification of artistic paintings.  
\end{itemize}

To explain how we achieve this, we first compare the Normalized Compression (NC), employing a data compression tool chosen after a competitive benchmark, with the Block Decomposition Method (BDM) \cite{Zenil-2018a}, and the inherent Coding Theorem Method (CTM) measures \cite{Delahaye-2012,Soler-2014a}. The BDM is an information-based measure that uses small Turing machines to approximate the algorithmic information, approximating to the Shannon entropy as a fallback mechanism. After this comparison, we make use of the average NC of each artist together with the roughness exponent $\alpha$ of the two-point height difference correlation function (HDC), to group artists by style. Furthermore, we provide a local complexity matrix that characterizes each artist using the NC and use it to construct a phylogenetic tree that portraits the relationship between artists in terms of exposing information to the observer. Finally, we use the regional complexity fingerprints and the roughness exponent $\alpha$ as useful auxiliary features that, combined with state-of-the-art approaches, improve the results of style and artist classification tasks.

The remaining of this paper is organized as follows. In the next section, we describe related work, followed by a description of the methods. We present the major results in the next section, with further results presented in Supplementary Material. Finally, we discuss the results obtained, draw final conclusions, and point out possible future lines of work.

\section*{Related Work}
Measuring the information contained in paintings requires fast, efficient, and automatic computation due to the diversity and large quantity of the existing artistic paintings \cite{Smiers-2003a}. To measure the information (or complexity) contained in paintings, we first need to define what is the quantity of information of an image. We define the quantity of information of an image as the smallest number of bits required by a model to represent an image losslessly. To perform this task, the model searches for unknown patterns of similarity between sub-regions of the image \cite{Ferreira-2014d,Pinho-2011c,Pratas-2012c} and uses this information to create this compressed representation of the image, relying exclusively in the patterns of the two-dimensional pixels without using exogenous information.

There are several approaches to quantify the amount of information. Kolmogorov described three, namely combinatorial \cite{Romashchenko-2002a,Niven-2009a,Mantaci-2008a}, probabilistic \cite{Shannon-1948a}, and algorithmic \cite{Kolmogorov-1965a}. Independently, the works of Solomonoff \cite{Solomonoff-1964a,Solomonoff-1964b} and Chaitin \cite{Chaitin-1966a} addressed the same lines. While the Kolmogorov complexity is non-computable, it can be approximated with programs for such purpose, such as data compressors, using probabilistic and algorithmic schemes.

Practical applications to approximate the Kolmogorov complexity for multiple dimensional digital objects have been developed using Turing machines \cite{Zenil-2018a,Soler-2017a,Gauvrit-2017a,Soler-2014a} and data compressors \cite{Ming-2001,Cilibrasi-2005a,Cilibrasi-2006,Cebrian-2007a,Cohen-2014,Pratas-2017c}. Recently, Zenil \textit{et al.} have shown that this methodology has a closer connection to algorithmic information than other measures based on statistical regularities \cite{Zenil-2018a}, namely fast lossless compression methods, for sources that follow algorithmic schemes.

The majority of the lossless compression algorithms are limited to finding simple statistical regularities as they have been designed for fast storage reduction \cite{Maniccam-2004,Lu-2016}; accordingly, they provide slight improvements over the Shannon entropy \cite{Shannon-1948a}. 
However, there are several which are designed for efficient compression at the expense of more computational resources. For example, lossless compression algorithms, such as GeCo \cite{Pratas-2016a}, are hybrids between probabilistic and algorithmic schemes. Besides having several context models of different orders, GeCo uses sub-programs that allow substitution \cite{Pratas-2017a} and reverse complement modeling \cite{Pinho-2008a}. These last two are sub-programs of probabilistic and algorithmic information nature. Another example is PAQ8 \cite{mahoney-web}, a general-purpose compressor that combines multiple context models using a neural network, transform functions, secondary structure predictors, and other simple sub-programs.
Usually, the problem is how to find fast and efficient algorithmic models for data compression. Lossless data compressors are tightly related to the concept of minimal description length \cite{Rissanen-1978a} and algorithmic probability \cite{Li-2004a,Cilibrasi-2005a,Li-2008b}. Therefore, representative algorithms can be efficiently embedded in these data compressors, including small Turing machines.

The idea of automatic computational analysis of artistic paintings is mature \cite{Lyu-2004a,Kim-2014a}, and the artistic community has widely relied on it for authentication of artistic paintings. Specifically, the characteristics of artistic paintings have been analysed through several statistical techniques and properties, namely fractal \cite{Taylor-1999a}, wavelet-based \cite{Lyu-2004a}, hidden Markov models \cite{Johnson-2008a,Li-2004d}, Fisher kernel based \cite{Bressan-2008a}, sparse coding model \cite{Olshausen-2010a,Hughes-2010b}, color and brightness \cite{Kim-2014a}, illumination \cite{Stork-2008a}, stroke \cite{Lettner-2008a,Shahram-2008a}, Print Index  \cite{Hedges-2008a}, and entropy-based analysis \cite{Petrov-2002a,Machado-2019a}. Recently, the work of Machado and Lopes \cite{Machado-2019a}, using fractional calculus, showed the potentiality of measures based on entropy to describe hierarchical clustering of paintings and their correlation with artistic movements. 

Regarding style and author classification, several recent works have proposed the usage of Convolutional Neural Networks (CNNs). A straightforward approach is to combine features extracted from multiple CNN layers, such as proposed by Peng \textit{et al.} \cite{peng-2015}.
Another more effective approach is based on representing images by the principal components of a Gram matrix that captures correlations across the different feature maps obtained from a convolutional layer of a pretrained deep CNN, such as VGG16 or VGG19. 
Mao \textit{et al.}~\cite{mao-2017}  combine this representation with the features from all the five convolutional blocks of the VGG16, learning a joint representation that can simultaneously capture content and style of visual arts. 
On the other hand, Chu \textit{et al.}~\cite{chu-2018} apply a support vector machine (SVM) to the Gram representation to perform author and style classification. Then, they improve the results by automatically learning correlations between feature maps.

\section*{Methods}
\label{Methods}
In this section, we describe the measures used, their normalizations, the methodology, and the compression benchmark performed.

\subsection{Information-based measures}
\label{probabilistic-algorithmic information measures}

Algorithmic information \cite{Solomonoff-1964a,Solomonoff-1964b,Kolmogorov-1965a,Chaitin-1966a} differs from a perspective of pure probabilistic information \cite{Shannon-1948a} because it considers that the source, rather than generating symbols from a probabilistic ergodic function, creates structures that represent algorithmic schemes \cite{Hammer-2000a,Henriques-2013a}. Therefore, to reverse the problem, there is the need to identify the program(s) and parameter(s) that generate the outcome(s) \cite{Kolmogorov-1965a,Chaitin-1966a,Li-2008b}. However, the algorithmic information, $K(x)$, is non-computable \cite{Terwijn-2011a}, mostly because of the halting problem \cite{Rybalov-2007a}. Therefore, we have to rely on approximations. Namely, in this subsection, we describe the Normalized Compression and two BDM normalizations. Then,  we establish the local application of the Normalized Compression to create a complexity matrix for each author and the methods used to create a distance matrix and the phylogenetic tree.
Finally, we describe a non-information-based measure, the two-point height difference correlation function.

\subsubsection*{Normalized Compression (NC)}

An efficient compressor, $C(x)$, gives a possible approximation for the Kolmogorov complexity ($K(x)$), where $K(x)<C(x)\le |x|$ ($|x|$ is the length of string $x$ in the appropriate scale). Usually, an efficient data compressor is a program that approximates both probabilistic and algorithmic sources using affordable computational resources (Time and RAM). Although the algorithmic nature may be more complex to model, data compressors may have embedded sub-programs to handle this nature. For a definition of safe approximation, see \cite{Bloem-2014a}. The normalized version, known as the Normalized Compression (NC), is defined by
\begin{equation}\label{NC}
\mathrm{NC}(x) = \frac{C(x)}{|x|\log_2{\mathrm{|A|}}} = \frac{C(x)}{|x|},
\end{equation}
where $x$ is a string, $C(x)$ is the compressed size of $x$ in bits, $|\mathrm{A}|$ the number of different elements in $x$ (size of the alphabet) and $|x|$ the length of $x$. Since we consider a binary matrix of each image, $|A|=2, \log_2{\mathrm{2}}=1$. Given the normalization, the NC enables to compare the information contained in the strings independently from their sizes \cite{Pratas-2017c}.  If the compressor is efficient, then the compressor can approximate the quantity of probabilistic-algorithmic information in data using affordable computational resources.

\subsubsection*{Normalized Block Decomposition Method (NBDM)}

Another possible approximation to the Kolmogorov complexity is given by the use of small Turing machines, where these small computer programs approximate the components of a broader representation. The Coding Theorem Method uses the algorithmic probability between a string's production frequency from a random program and its algorithmic complexity.  As such, the more frequent a string is, the lower Kolmogorov complexity it has; and strings of lower frequency have higher Kolmogorov complexity. The Block Decomposition Method (BDM) extends the power of a CTM, approximating local estimations of algorithmic information based on Solomonoff-Levin's algorithmic probability theory. In practice, it approximates the algorithmic information and, when it loses accuracy, it approximates the Shannon entropy.
Since in this article we intend to perform a direct comparison of both measures, we first considered the normalization of the BDM (NBDM$_1$), given by the number of elements (length) of the digital object as

\begin{equation}\label{NBDM_1}
\mathrm{NBDM_1}(x) = \frac{BDM(x)}{|x|\log_2{\mathrm{|A|}}} = \frac{BDM(x)}{|x|}.
\end{equation}

However, the normalization of the BDM is usually performed using a minimum complexity object (BDM$_{Min}$) and a maximum complexity object (BDM$_{Max}$).
A minimum complexity object is filled with only one symbol, like a binary string of only zeros. In contrast, a maximum complexity object is an object that, when decomposed (by a given decomposition algorithm), yields slices that cover the highest CTM values and are repeated only after all possible slices of a given shape have been used once.
Using these two objects, the NBDM$_2$ for a given string can be computed as
\begin{equation}
\label{NBDM_2}
\mathrm{NBDM_2}(x) = \frac{BDM(x) - BDM_{Min}}{BDM_{Max} - BDM_{Min}},
\end{equation}

where $BDM(x)$ is the BDM value of that string, $BDM_{Min}$ is the minimum complexity object, and $BDM_{Max}$ is the maximum complexity object.

Kolmogorov complexity is invariant only up to a constant factor, which depends on the choice of a description language \(K = K' + L\), where $K$ is the total complexity, $K'$ is the description of the object and $L$ is the description of the language. As such, by performing the normalization according to Equation \ref{NBDM_2}, the normalization is aiming to remove the constant factor as

\begin{equation}
\label{constant_factor}
\frac{K - K_{Min} }{K_{Max} - K_{Min}} = \frac{K' + L - K'_{Min} - L }{K'_{Max} + L - K'_{Min} - L} = \frac{K' - K'_{Min} }{K'_{Max} - K'_{Min}},
\end{equation}
where $K_{Max}$ and $K_{Min}$ are the maximum and minimum Kolmogorov complexity objects and $K'_{Max}$ and $K'_{Min}$ are the maximum and minimum Kolmogorov complexity description of the objects.

In this article, we perform a direct comparison between the NC and the NBDM$_{1}$. Furthermore, we compare the two types of BDM normalization and their impact on the results.

\subsubsection*{Local complexity analysis using the Normalized Compression}
\label{region}

The Normalized Compression (NC) was used to approximate the local (or regional) complexity of images of artistic paintings. To that end, all of the dataset images were divided into 16x16 blocks (256 equal regions) and the NC was computed for each block, generating a complexity matrix. Other patch sizes were also tested, specifically patch sizes of 8x8 and 32x32 blocks.
Following this operation, the average complexity matrix was generated for each author, using the complexity matrices of their paintings. The average complexity matrices were then used to obtain a similarity matrix, in which the distance between matrices was determined as

\begin{equation}
\label{sim}
\mathrm{d(A,B)} = \sum_{i=0}^{n} \sum_{j=0}^{n}{ \mid a_{ij}-b_{ij}\mid },
\end{equation}

where $d$ is the distance between the complexity matrix $A$ and $B$, and $a_{ij}$ and $b_{ij}$ are the complexity values at the index $i$ and $j$ of matrices $A$ and $B$, respectively.
Subsequently, using the similarity matrix, a phylogenetic tree was computed recurring to two methods, namely UPGMA (unweighted pair group method with arithmetic mean) \cite{Sokal-1958a} and the Kruskal minimum spanning tree algorithm \cite{Kruskal-1956}, in order to portrait complexity relationships among different authors. 

\subsection{Two-point height difference correlation function}
The two-point height difference correlation (HDC) function was computed to quantify brightness contrast as

\begin{equation}
\label{HDC}
\mathrm{HDC}(r) = \overline{[h(\vec{x} + \vec{r}) - h(\vec{x})]^2} = \frac{1}{N_{r}}\sum_{\vec{x},|\vec{r}|=r}[h(\vec{x} + \vec{r}) - h(\vec{x})]^2,
\end{equation}

where the $r$ is the distance between two-pixel points, over-bar represents the spatial average at a fixed distance $r$ for all possible points; $N_r$ is the number of possible pairs at a distance $r$, $h(x)$ is pixel intensity at the position $x$. Using the HDC function, its roughness exponent $\alpha$ was determined as
\begin{equation}
\label{alpha}
\mathrm{\alpha} = \frac{\log_{10}(HDC(r_{final})) - \log_{10}(HDC(r_{initial}))}{\log_{10}(r_{final}) - \log_{10}(r_{initial})},
\end{equation}

where ($\alpha$) is the slope of the HDC curve in a double logarithmic plot of the surface growth model. The slope was calculated from $r_{initial} = 10$ to $r_{final}$, which matches the point where the HDC function saturates, approximately 30\% of the image's width.

\subsubsection*{Assessment pipeline}
\label{Assessment Pipeline}
In order to fairly evaluate the information-based measures, we designed a pipeline for processing images. It respects the following steps: Obtaining the dataset images;  converting the images to PGM format; quantization of the images to 8 bits (256 levels) using the Lloyd-Max algorithm; binarization of the images (conversion to 01 format in ASCII) and finally, applying the information-based measurements (NC, NBDM$_1$ and NBDM$_2$).

Quantization was performed using the Lloyd-Max algorithm \cite{Lloyd-1982a,Taubman-2002a} since reducing the precision of the pixels (alphabet) in images enables the filtering of small variations that might occur during the digitalization process.
Binarization to 01 format in ASCII was performed since the BDM currently only supports a small alphabet. 

\subsubsection*{Finding an effective data compressor}
\label{Finding an effective data compressor}

To compute the NC, we have to find an effective data compressor, meaning, a compressor that best represents each image, while using reasonable resources. Since our aim is later to apply this measure to a dataset of artistic painting, we compared seven compression tools, namely GZIP \cite{gzip-web}, BZIP2 \cite{bzip2-web}, XZ \cite{xz-web}, LZMA \cite{7zip-web}, AC \cite{hosseini2019ac}, PPMD \cite{Cleary-1984a}, and PAQ8 \cite{mahoney-web}.

\begin{figure}[ht!]
\centering
\includegraphics[width=8.6cm]{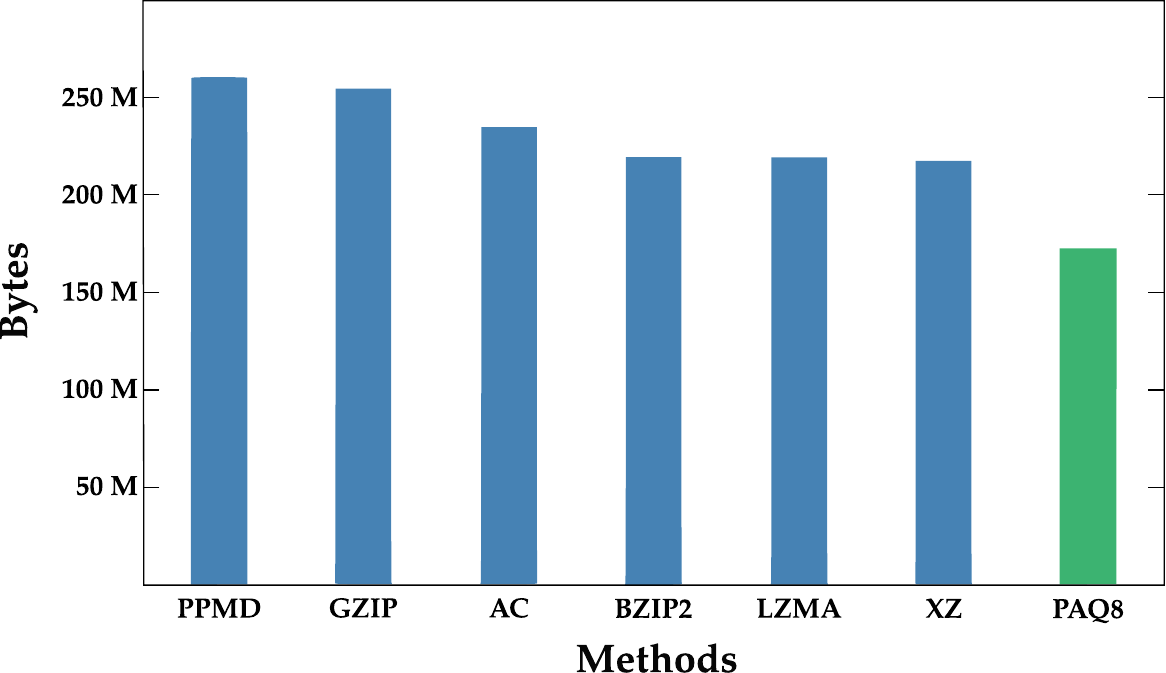}
\caption{Benchmark of lossless data compression tools specifically for the processed dataset of artistic paintings. The y-axis depicts the sum of the number of bytes to compress the dataset, where each image was compressed individually using each tool}
\label{bench}
\end{figure}

As depicted in Figure~\ref{bench}, the PAQ8 tool shows the best compression ratio for this dataset. In fact, it shows an improvement of $\approx 26\%$ to the second best tool (XZ). The disadvantage is the use of higher RAM and substantially more computational time. Nevertheless, since our purpose is to find the number of bits of a shortest program to reproduce the image, it is affordable to spend these computational resources. Therefore, we used the PAQ8 tool to compress each of the quantized images. The code was compiled using the package provided from \cite{paq-web}. The PAQ8 version used was kx v7. PAQ8kx v7 is an archiver that achieves the highest compression rates at the expense of speed and memory (approximately 1,6 GB of RAM for this dataset). We used the mode that usually provides the highest compression ratio (command parameter: ``-8''). The PAQ8 compressor uses a context mixing algorithm between a large number of models independently predicting each quantized pixel's next bit \cite{Mahoney-2005a}. The predictions are combined using a neural network and arithmetic coding \cite{Rissanen-1979a,Moffat-1998a}. For automatic installation, use the script \codeword{Install.sh}, while for more information of PAQ, see the work of Knoll and Freitas \cite{Knoll-2012a}. The computations ran in a single core Ubuntu Linux computer running at 2.13 GHz with 1.6 GB of RAM. Using this machine, the compression of the whole dataset with PAQ8 required approximately 270 hours of real-time, without parallelization.

\section*{Results}
\subsection*{Comparison of NC and BDM}

In order to compare NC with BDM, we performed three types of tests. Namely, we compared the robustness of both measures according to increasing rates of random pixel changes in paintings, tested their application on different types of images, and made an assessment of the minimal information bounds.

In the first test, we assessed the impact of an increasing rate of pixel editions using a pseudo-random uniform distribution and compared both information-based measures. This approach is not identical to image noise, but rather a pure edition of pixels. For the purpose, for each of the three authors (Theodore Gericault, Marc Chagall, and Rene Magritte) we select a painting, making 50 adulterated copies of each painting with increasing edition rate (from 1 to 50\%). Finally, we measured the NC (Eq.~\ref{NC}), the NBDM$_1$ (Eq.~\ref{NBDM_1}), and NBDM$_2$ (Eq.~\ref{NBDM_2}) in all the paintings.

\begin{figure*}[ht!]%
\centering
\includegraphics[width=\textwidth]{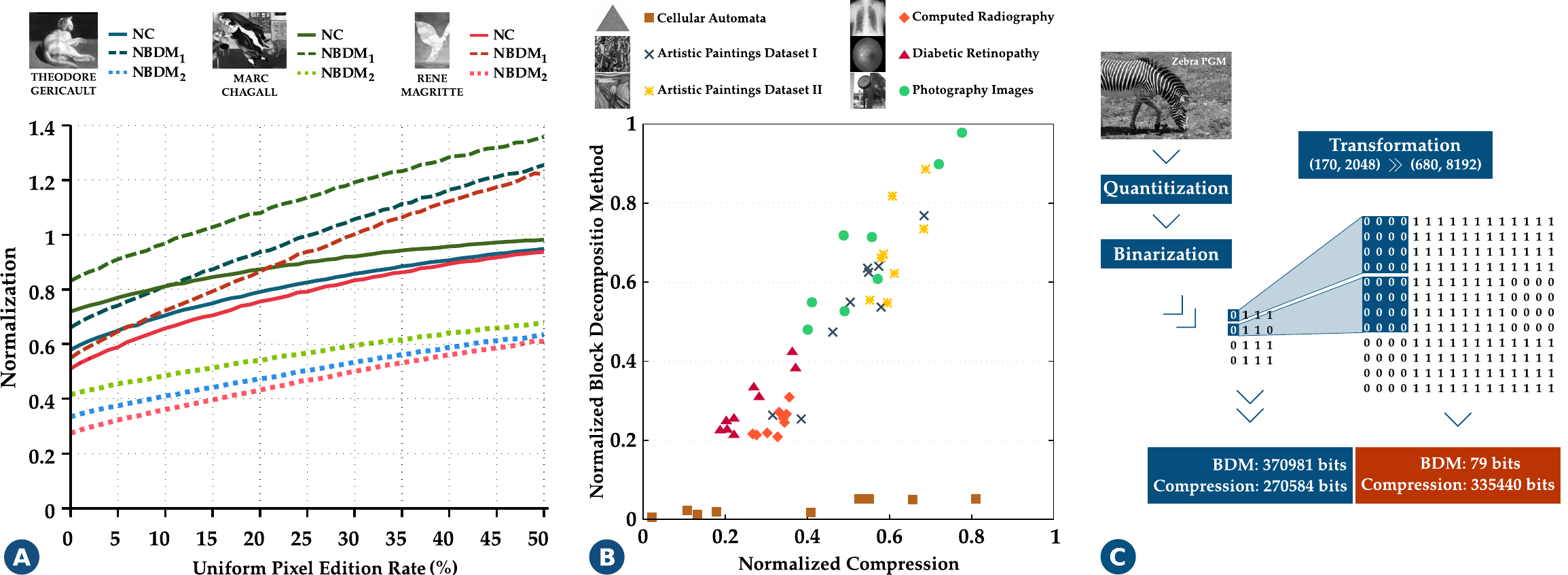}
\caption{Information-based measures evaluation. (\textbf{A}) Impact of increasing pseudo-random substitution on information-based measures: NC (approximated using the PAQ8 algorithm) and two BDM normalizations (NBDM$_1$ and NBDM$_2$). (\textbf{B}) Values of the NC and NBDM$_1$ for different types of images.  (\textbf{C}) Image transformation pipeline leading to BDM underestimation of the amount of information contained in the transformed object.}
\label{nc_nbdm1_nbdm2}
\end{figure*}

Figure~\ref{nc_nbdm1_nbdm2} (\textbf{A}) depicts the values obtained for the NC and BDM. The results show that, when using the same type of normalization, NC is more robust to the increment of pixel edition than NBDM (NBDM$_1$). On the other hand, whereas NBDM$_1$ considers the normalization by the length of the input object, NBDM$_2$ performs a normalization that aims to mimic the removal of the constant factor related to Kolmogorov complexity (see Eq. \ref{constant_factor}). Since the NBDM$_2$ normalization does not take into account the constant of the description language, it shows a more robust behavior than NBDM$_1$, which increases rapidly with the increase of pixel edition.
Since NC and NBDM$_1$ have the same type of normalization, we will focus on comparing these normalizations from now on.

In the second test, we applied both measures to six datasets with distinct nature (9 images each) to understand how NBDM$_1$ and NC behave with different types of images. The six datasets were: artistic images from 2 different datasets \cite{Khan-2014a,art-web}; cellular automata images; diabetic retinopathy images \cite{DR-web}; chest computed radiography (CR) images \cite{wang-2017} and  photographic images \cite{coco-web}.

The results are depicted in Figure~\ref{nc_nbdm1_nbdm2} (\textbf{B}). Overall, the majority of the datasets show similar behavior regarding the NC and NBDM$_1$. The exceptions to this are the CR and cellular automata datasets, which exhibit a more algorithmic behavior.
The latter dataset is constituted by images created with small programs with simple rules.  Whereas the compressor has difficulty compressing this type of images, the BDM can point to their algorithmic nature, and, thus attribute them with minimal value. This outcome shows the importance of the BDM in the detection of simple algorithmic outputs embedded into data.

In the last test, we selected one of the most complex images identified by the NBDM in the last subsection to test if the BDM could accommodate specific data alterations. This test is depicted in Figure~\ref{nc_nbdm1_nbdm2} (\textbf{C}). After the binarization process, we performed a super-sample image transformation where each char was amplified to a 4x4 representation. This value was selected since the BDM has the default block size value of 4x4 in 2D structures.
After this operation, the BDM was computed for the original and the super-sampled image. While the original image was measured with 370981 bits, the super-sampled image had only 79 bits. This abrupt decrease in the complexity value indicates that the BDM underestimates the amount of information contained in the object. The BDM analyses object information in blocks instead of looking at the whole object. Specifically, blocks analysed by the BDM (default block size value of 4x4 in 2D structures) have the same size as the super-sample image transformation (each char was amplified to a 4x4 representation); therefore, the complexity attributed to each block is approximately zero (since each block is composed of all zeros or ones), and hence the overall value attributed to the complexity of the object will drop dramatically. 

This analysis shows that BDM is not prepared to deal with the information associated with the choice of the model, unlike the NC. The NC relies on the use of a lossless data compressor, bounded by a maximum information channel capacity. 

From these three tests, we are able to notice some advantages and limitations of both measures. Ranking these measures is not a fair task because they have different characteristics and nature. Therefore, in the remainder of the article, we use the NC and NBDM in a combined mode to recover insights and characteristics from the images of the artistic paintings.

\subsection*{Information-based measures in images of artistic paintings}

Herein, we investigate the use of information measures to analyse a dataset of artistic paintings. This dataset \cite{Khan-2014a} contains 4,266 images of artistic paintings from 91 authors, with approximate geometric sizes. The 91 authors are well-known painters, such as Claude Monet, Frida Kahlo, Henri Matisse, Jackson Pollock, Picasso, Rembrandt, and Salvador Dali. 
In the following subsections, we present the results of applying the measures, combining the NC with the HDC function, measuring local complexity for different authors and constructing a phylogenetic tree, as well as using these features to improve style and artist classification.

We also measure the impact of normalizing these images by performing image normalization and then applying the measures mentioned above in the dataset. Afterwards, we compared the average variation difference and the percentage difference between the results obtained for each author. The results are shown in the Supplementary Material in section \ref{normalization_analysis}.

\subsubsection*{Global measures analysis}
\label{Probabilistic-Algorithmic Analysis}

In this subsection, we measure an approximation to the Kolmogorov complexity for the dataset of artistic paintings. The same pipeline, described in the methods section, was used, with the difference that the Lloyd-Max algorithm quantization was set to 16, 64, and 256 levels (4, 6, and 8 bits respectively). Important to note that Lloyd-Max algorithm forced normalization of the images for the 16 and 64 levels, while the 256 level was the original level of the images, and, as such, these images were not normalized. This process was performed to evaluate the impact of the quantization on the measures used to approximate the Kolmogorov complexity in artistic painting images. From the results obtained from the measures, we show unknown characteristics and insights into temporal traits.

\begin{figure*}[ht!]
\centering
\includegraphics[width=\textwidth]{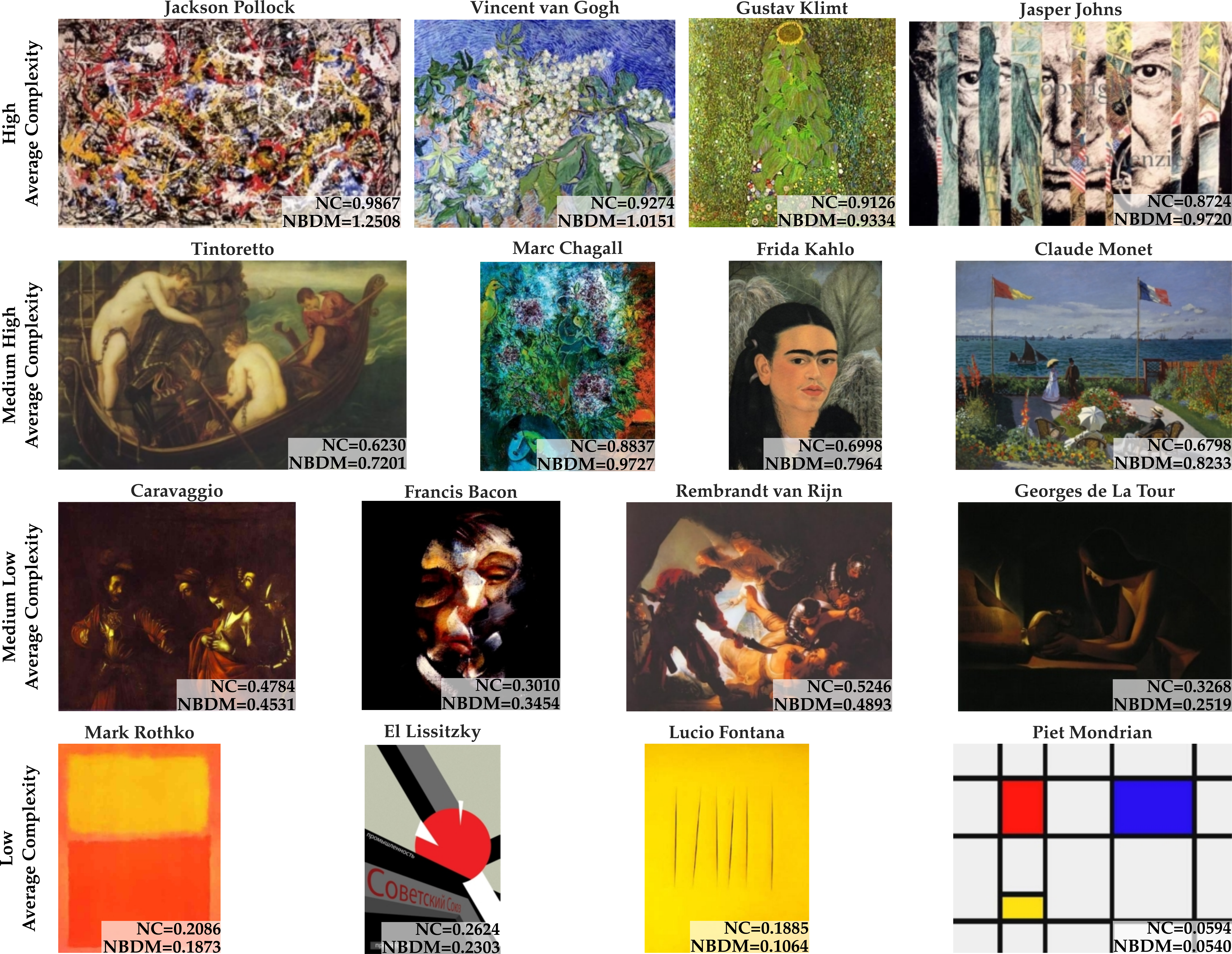}
\caption{Examples of artistic paintings with different levels of complexity where painting images were quantized to 8 bits. The NC and NBDM$_1$ values of each painting are displayed in its lower right corner.}
\label{Artists_paintings}
\end{figure*}

In general, the complexity of each painting follows the example of Figure~\ref{Artists_paintings}. Paintings with low complexity are classified as abstract and minimalist, following simple patterns. As the complexity increases, we start to recognize paintings with different local complexities, meaning, there are regions with high complexity and detail (generally on the center/bottom of the paintings) surrounded by low complexity regions (same color background) namely known as chiaroscuro. This pattern begins fading, as the complexity increases since the highest complexity paintings are also the most irregular, detailed, and convoluted.

\begin{figure*}[ht!]%
\centering
\includegraphics[width=\textwidth]{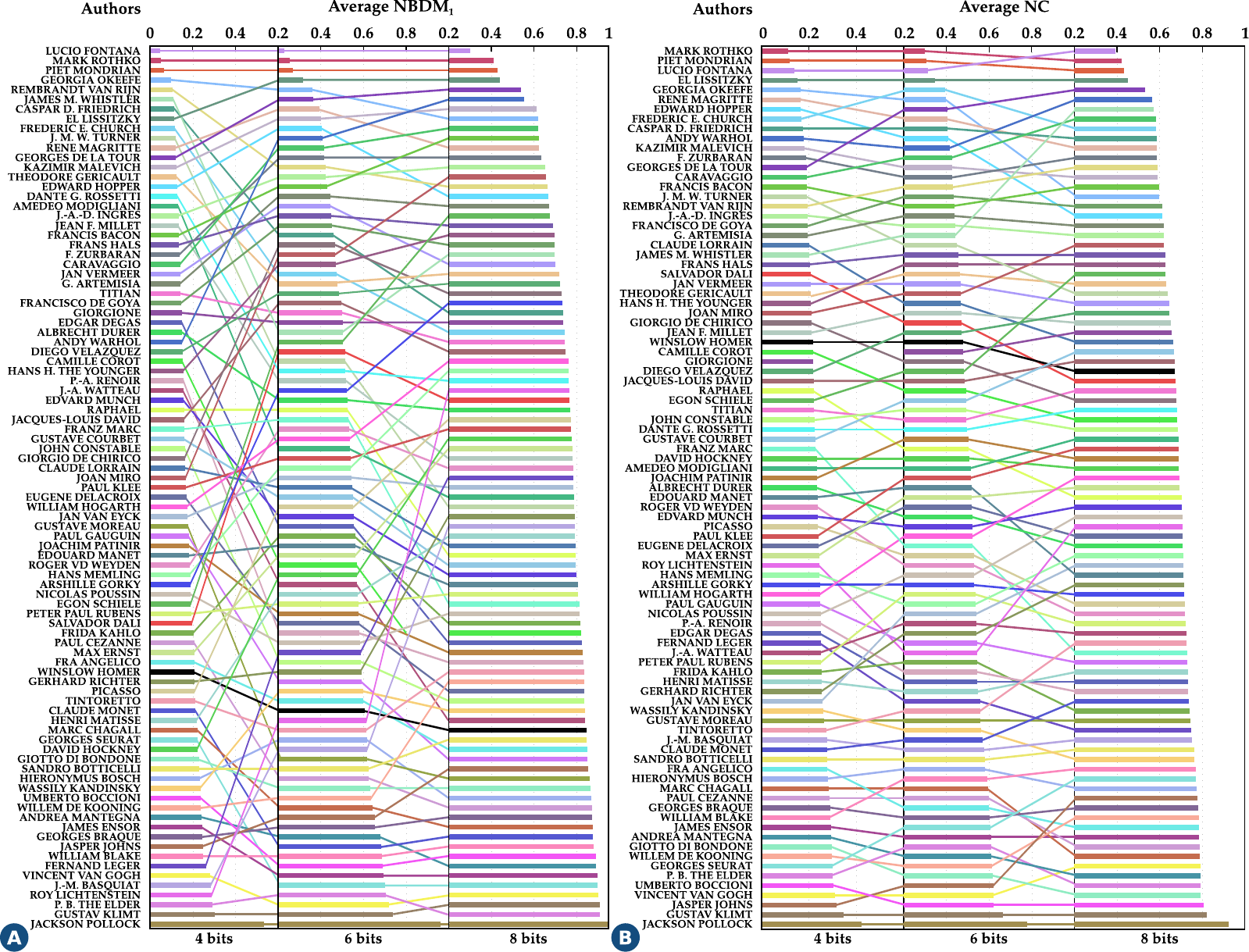}
\caption{Average Normalized Block Decomposition Method using NBDM$_1$ (\textbf{A}), and Average Normalized Compression (\textbf{B}) for each author where images of paintings where quantized for 4, 6, and 8 bits. The authors are sort given the value of NBDM$_1$ and NC, respectively. To see this result in more detail, please visit the website associated with the article.}
\label{Average_complexity}
\end{figure*}

Regarding the average complexity values for each artist, Figure~\ref{Average_complexity} shows the average of NBDM$_1$ and NC, respectively. Each artist has an associated color, and lines of the same color illustrate its relative positional deviation in different quantizations.
The same results for NBDM$_2$ are exposed and discussed in the Supplementary Material.
Noticeably, quantization impacts the NBDM$_1$  more than the NC, since the relative positioning between authors varies more in the former. On average, the variation is $13.4 \pm 11.37$ relative positions of each author in NBDM$_1$, while in NC, the variation is $4.9 \pm 4.3$ positions.

Despite the higher variation present in the NBDM$_1$, both measures are capable of detecting styles with low and high complexity. Artists such as Mark Rothko, Lucio Fontana, Piet Mondrian, El Lissitzky can be easily identified on the low side of the complexity spectrum.  Minimalism, Abstract Expressionism, and Constructivism movements are associated with these styles.
On the other hand, artists from Abstract Expressionism, such as  Willem de Kooning, Jackson Pollock, and Jasper Johns, characterize the highest complexity side of the spectrum, as well as other artists with a more detailed and convoluted style, like Gustav Klimt and Vincent van Gogh.

Abstract Expressionism is characterized by aggressive features combined with random and geometric features and spontaneity \cite{shapiro-1978}.
The reason for Abstract Expressionism artists being present at both extremes of the complexity spectrum is because this style itself divided into two opposites, Action Painting and Color Field.
In Action Painting, the paint was thrown directly on the canvas, through instinctive gestures, where chance and randomness determined the evolution of painting \cite{rosenberg-1994}. This style is characteristic of artists like Jackson Pollock (known for the technique of “dripping”) and Willem de Kooning. On the other hand, Color Field is more mystical and meditative. This style of painting has few elements in the frames, indefinite limits, and explores the sensory effects of color, as well as the subtlety of chromatic relations \cite{garrard-2007}. A specific example of an artist that followed this trend was Mark Rothko.
In all cases, Jackson Pollock had complexity values utterly different from other artists, the average complexity of his paintings being approximate to random (normalized value close to 1). Although he denied his paintings were random, similar results were also found in previous work, which defined Jackson Pollock's dripping paintings as not typical artworks~\cite{Kim-2014a}.

\subsubsection*{Combining the NC with the roughness exponent of HDC function}

\begin{figure*}[ht!]%
\centering
\includegraphics[width=\textwidth]{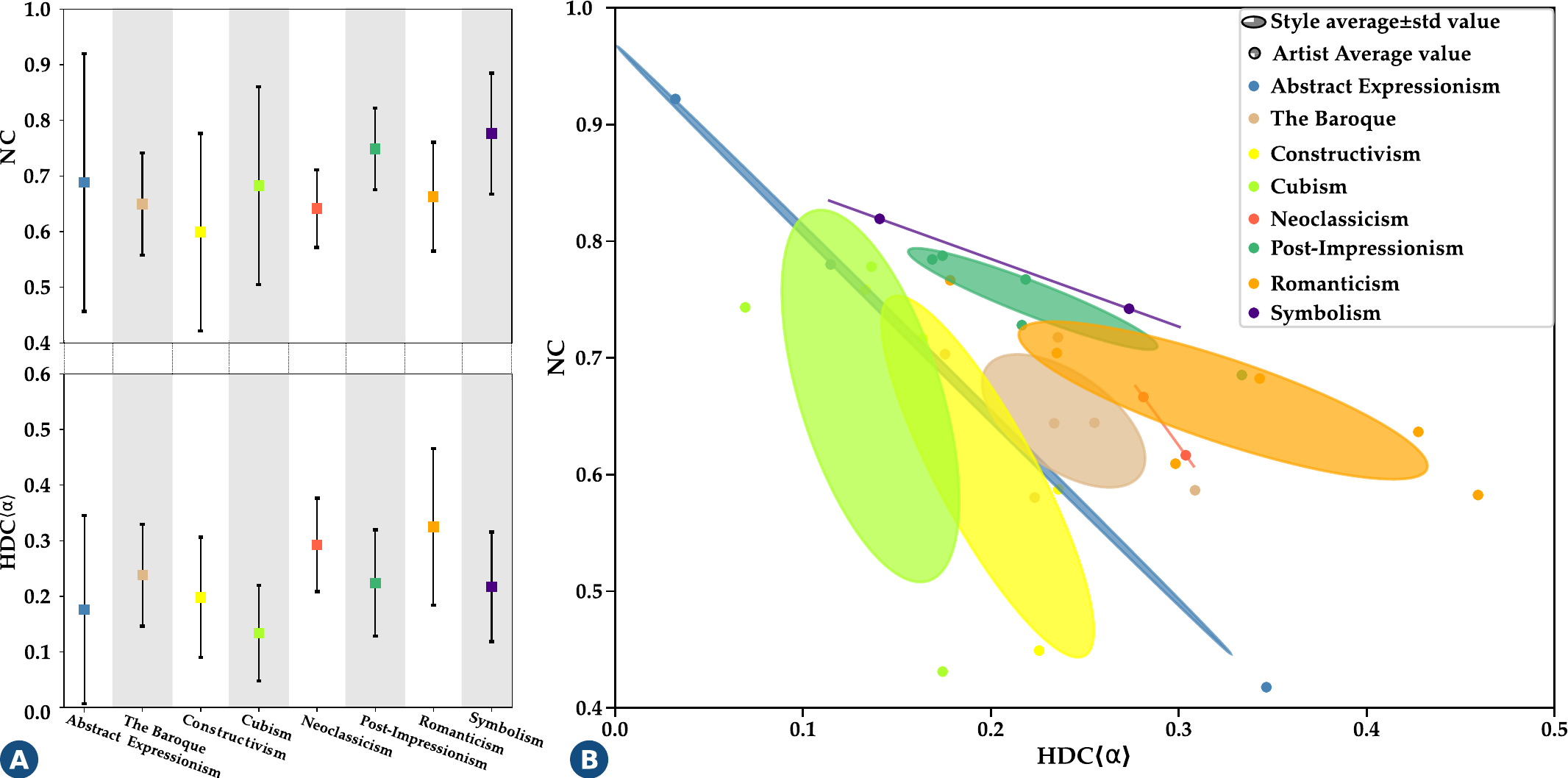}
\caption{Combining the HDC with NC. (\textbf{A}) Average and standard deviation for each style in NC and $\alpha$, respectively. (\textbf{B}) Results grouped by styles using average NC and average $\alpha$ of HDC for each artist labeled on the dataset.}
\label{hdc_nc}
\end{figure*}

We used the average NC together with the roughness exponent ($\alpha$) of the two-point height difference correlation (HDC) function, which measures the roughness exponents of brightness surfaces, to assess the ability of these measures to distinguish different styles. Accordingly, we made usage of style labeled paintings available in the dataset. From these labeled images, we computed for its author the average NC and the value of $\alpha$. The roughness exponent was used as an additional measure since it has proven to be capable of some differentiation between styles\cite{Kim-2014a}. We discarded the usage of BDM due to quantization impacting it more than the NC.
Using the average NC and $\alpha$ of each labeled painter, we created a scatter plot (Figure~\ref{hdc_nc}) and represented each artistic movement as an ellipse, with the center in the points' center of mass and with a width corresponding to the standard deviation. 

As shown in Figure~\ref{hdc_nc} (\textbf{A}), both measures alone are not capable of efficiently separating styles, However, when combined, the styles are well confined into different regions (except for Abstract Expressionism), showing that together these measures are representatives of artistic movements. The roughness exponent $\alpha$ captures the level of brightness and relative spatial position and is correlated to variations in painting techniques and genres \cite{Kim-2014a}. The NC adds to the level of brightness and relative spatial position provided by the HDC, the notion of average information present in each artist's painting. This amount of information differs depending on the artistic movement and historical circumstances.

Interestingly, similar to NC, the roughness exponent of the HDC varies greatly in Abstract Expressionism, being that in this artistic movement, there is an inverse correlation between the NC and $\alpha$. Namely, artists like Jackson Pollock and Willem de Kooning (Action Painting) presented a high average NC and a low $\alpha$, whereas, Mark Rothko (Color Field) had polar results.  This atypical behavior corroborates the big difference between the two currents of Abstract Expressionism. The Action Painting usage of instinctive gestures and randomness creates high NC values and spatial correlation approaching a random image. In contrast, in Color Field, we get more minimalist images with high spatial contrast between regions but low complexity.

\subsubsection*{Local complexity of paintings}

\begin{figure*}[ht!]%
\centering
\includegraphics[width=.99\textwidth]{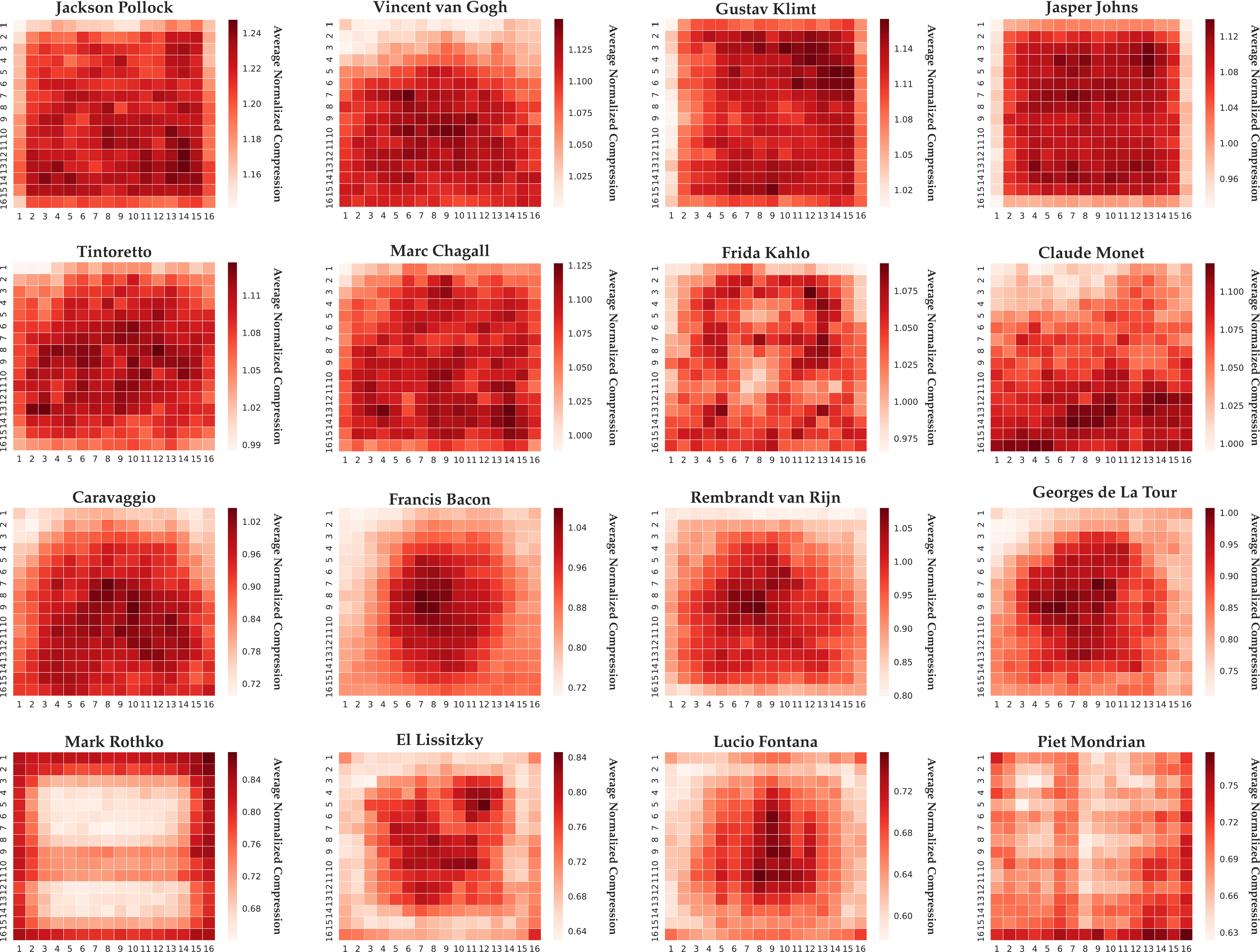}
\caption{Heat maps of the local complexity matrix (fingerprint) of some authors, computed with the NC. This fingerprint shows the author's range of complexity and the locations in the canvas painted with more detail (or complexity). To see all matrices, please visit the website associated with this article.}
\label{example_fingerprint}
\end{figure*}

In this section, we divided the images into identical quadrilateral sizes and measure the algorithmic information for each one (16x16 blocks). Then, we computed the average of each quadrilateral for all the paintings for each painter. 
The results are shown in Figure~\ref{example_fingerprint}, illustrating the same authors as those in Figure~\ref{Artists_paintings}. Note however that matrices of Figure~\ref{example_fingerprint} were computed using all the authors' paintings present in the dataset. The complete results are available on the website associated with this article.
The same computation was repeated for blocks of sizes 8x8 and 32x32. Analysis of these results, included in the Supplementary Material, show that 16x16 is the minimum patch size for which the differences in the compression rate are noticeable and can therefore be used as a measure between paintings.

All artists have a unique complexity matrix (fingerprint). This fingerprint shows, on average, where artists paint with more detail and give more emphasis as well as the average range of complexity the artist operates. For instance, Jackson Pollock and Jasper Johns show high complexity values dispersed over the canvas. At the same time, artists like Francis Bacon and George de la Tour focus more on the center of the canvas, and Mark Rothko and Piet Mondrian give their highest complexities around the borders of paintings.

\begin{figure*}[ht!]%
\centering
\includegraphics[width=1\textwidth]{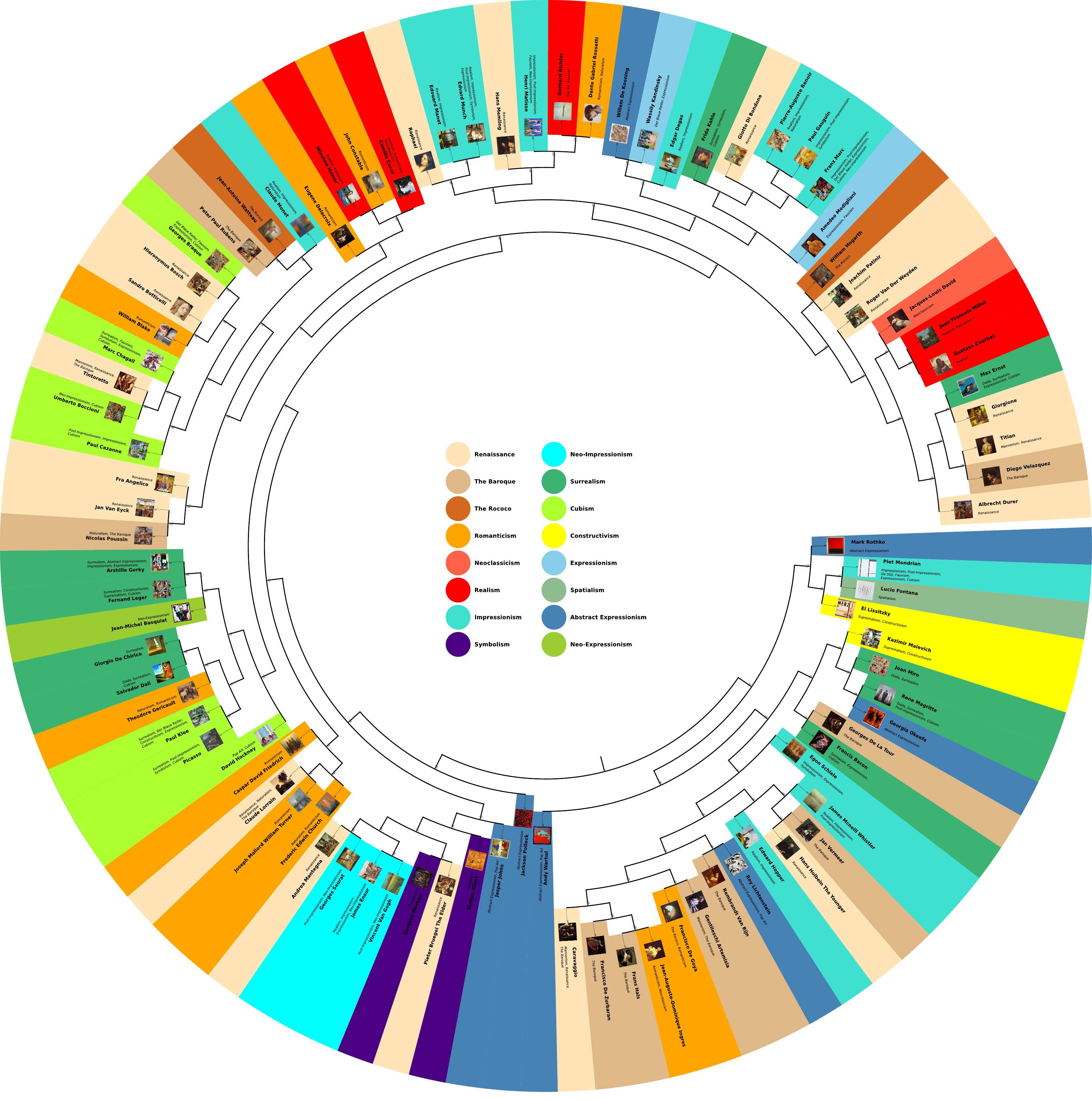}
\caption{Artists' phylogenetic tree computed recurring to the UPGMA algorithm. Each artist has a sample painting and a colour associated with one of his styles (the colour was chosen based on nearest leaves) assigned to him, as well as a description of some styles usually associated with the author. To obtain an improved view of the tree, please visit the website related to this article.} 
\label{tree_upgma}
\end{figure*}

Since the 16x16 fingerprints conveyed the best results regarding detail and differentiation (see Supplementary Material in section \ref{fingerprint_comparison_section}), the phylogenetic trees were constructed utilizing the distance computed from the fingerprints with block size. Concretely, two phylogenetic trees were constructed to portray the relations between different artists. One tree was constructed using the UPGMA algorithm, which is illustrated in Figure~\ref{tree_upgma}, and another tree was build using the Kruskal minimum spanning tree algorithm \cite{Kruskal-1956}, which is depicted in the Figure \ref{Kruskal_tree} of the Supplementary Material in section \ref{Kruskal_section}.

The tree shows the fingerprint's capacity of grouping artists from the same artistic movements mutually. Broad groupings of artists from styles are present in the tree, namely, Renaissance, Baroque, Romanticists, Impressionists, Surrealism, Cubism, and Abstract Expressionism. Also, the tree shows smaller groupings of sister leaf-nodes with the same style.
On the other hand, the tree depicts relationships of influence between authors of different artistic movements. This relation is seen in the case of Titian, who influenced Diego Velazquez; Caravaggio, who influenced Francisco de Zurbar\'{a}n; Frida Kahlo, who influenced Amedeo Modigliani; Sandro Botticelli who influenced William Blake; Claude Lorrain who influenced Joseph Mallord William Turner; and Peter Paul Rubens who influenced Jean-Antoine Watteau.

On the other hand, some authors seem unrelated in style and influence, for instance, Francis Bacon and Georges de la Tour, George Braque and Hieronymus Bosch, Peter Paul Rubens and Frida Kahlo, Max Ernst and Giorgione, and Rembrandt van Rijn and Roy Lichtenstein. There can be many reasons for this to occur, for instance, the number of regions the images were divided can be sub-optimal for some images of artistic paintings, decreasing the sensitivity of the measure and jeopardizing the tree's construction. On the other hand, the algorithm used to measure the similarity between matrices or the algorithm used to construct the tree (UPGMA) may not be the most appropriate for all cases, however, we have tested the Kruskal minimum spanning tree algorithm which yielded similar results (see Supplementary Material in section \ref{Kruskal_section}).
Additionally, these seemingly unrelated connections could reveal undiscovered elements and relationships. For instance, one of Roy Lichtenstein's early artistic idols was Rembrandt van Rijn. 
Moreover, if artists are not related regarding the artistic movement or influence, the vicinity among them could be representing another property. This aspect is not necessarily related to the period or movement the artists were inserted in, but rather, the way authors projected their compositions, ideas, and impressions onto the canvas. Complexity can be approximated by the total number of properties transmitted by an object and detected by an observer. By dividing images into blocks of equal size and evaluating its local complexity, we are quantifying the local information being transmitted. On the other hand, by averaging the canvas results per artist, we obtain a matrix that describes how the author exposes information to the observer.
This information intertwines various notions critical to how the work is perceived, such as composition which describes where the artist places the subject and how the background elements support it, as well as the unity, balance, movement, rhythm, focus, contrast, pattern, and proportion of the painting. For instance, the proximity between Hans Holbein and Vermeer could be due to both of them having used optics to achieve precise positioning in their compositions, namely by performing a combination of curved mirrors, \textit{camera obscura}, and \textit{camera lucida} \cite{hockney-2006}.

Another example that this information can convey is space by depicting where positive (subject itself, which is usually more detailed) and negative (the area of painting around it) spaces are on the canvas. Artists can play with a balance between these two spaces to further influence how viewers interpret their work. Therefore, the similarity between different artists concerning the regional (local) complexity can reflect the similarity in thought regarding their approaches to painting. 
For instance, the proximity between Francis Bacon and Georges de la Tour could be due to the former being heavily influenced by the Baroque style and having made dramatic use of contrasts of light and shadow. These methods are characteristic of the \textit{chiaroscuro} principle and its radicalization in the Tenebrista school (signature style of Georges de la Tour) \cite{Yang-2016,Fichner-2011}. The intense contrasts of light and shadow highlight the characters, and although exaggerated, it is lighting that increases the feeling of realism, making the muscles and facial expressions more evident. Simultaneously, the presence of large blackened areas highlights the chromatic research and the illuminated space, which acquire their value as elements of the composition.

We conclude that this novel technique is a unique descriptor of the authors' paintings since it not only aggregates authors of the same style close to each other and demonstrates the influences that authors had on others. It also serves as an insight into the way the artist projects its art. 
 
\subsubsection*{Evaluation of measures for classification purposes}
 
To quantitatively evaluate the use of these measures for classification purposes, we assessed their impact when used as additional features to improve state-of-the-art classification methods.
For this purpose, we recreated a recently published state-of-the-art method as a baseline and improved the results by combining our proposed measures.
Based on current methods~\cite{mao-2017,chu-2018}, we extracted a Gram representation using the first convolutional layer from the fifth convolutional block of the VGG16 network which was pre-trained with the imagenet dataset (no significant result difference was found between the use of VGG16 or VGG19). Principal component analysis (PCA) was applied to the Gram matrix to reduce the dimensionality and, finally, this vector was provided to an SVM to perform classification.

Afterwards, the features obtained from computing the HDC and the regional complexity were used for author and style classification using the XGBoost classifier \cite{Chen-2016} and combined with the baseline classifier via a Voting Classifier ensemble. 
The results of the baseline and ensemble classifiers, applied to the Paintings91 dataset in the author and style classification task using the labels provided in the dataset, are shown in Table~\ref{tab:Classification}.

\begin{table}[ht]
\setlength{\tabcolsep}{6pt}

\caption{Accuracy results obtained for the test set in style and author classification task using state-of-the-art (SoA), state-of-the-art with regional complexity (RC) and ensemble with our measures (RC and HDC).}
\centering
\label{tab:Classification}
\smallskip
\begin{tabular}{T{0.18\textwidth}T{0.10\textwidth}T{0.10\textwidth}T{0.12\textwidth}T{0.12\textwidth}T{0.13\textwidth}}
\toprule
\textbf{Classification Task} & \textbf{Number of Classes} & \textbf{Number of Images} & \textbf{SoA Baseline} & \textbf{SoA Baseline + RC}  & \textbf{SoA Baseline + RC + HDC}\\
\midrule
\textbf{Style} & 13 & 2338 & 0.622 & 0.644 & 0.650\\
\textbf{Author} & 91 & 4266 & 0.480 & 0.490 & 0.500\\
\bottomrule
\end{tabular}
\end{table}

The results show that the inclusion of the Regional Complexity, increased the accuracy of the results 2.2 p.p. and 1.0 p.p in the style and author classification tasks respectively. Moreover, the overall inclusion of the proposed measures (HDC + RC) increased the accuracy in both classification tasks by 2.8 p.p. and, 2.0 p.p. in the style and author classification tasks, respectively. These results indicate that these predictors are useful auxiliary features capable of improving current methodologies in the classification of artistic paintings.
This is congruent with the results obtained with Nanny \textit{et al.} \cite{Nanni-2017}, since handcrafted features and non-handcrafted features seem to extract different information from the input images and, as a result, the fusion of the two types of features improves the results obtained when using non-handcrafted features only.
Furthermore, regional complexity (RC) has a higher impact on the improvement of the accuracy than the HDC features, demonstrating the importance and distinction of Regional Complexity as a feature.

\section*{Discussion}

In this work, we develop, use, and compare unsupervised pattern recognition techniques to quantify information in images of artistic paintings. We rely on two approaches, namely data compression using the Normalized Compression (NC), and the Block Decomposition Method (BDM), to estimate information of both probabilistic and algorithmic sources. To approximate the NC, we benchmark a set of data compressors, where we show that the most effective for this dataset is PAQ8. Subsequently, this article is organized into two broad sections. The first is the evaluation and comparison of information-based measures; the second is applying these information-based measures to a dataset of artistic paintings. 

On the measure evaluation section, we assessed the NC and BDM using three tests. In the first test, we evaluated the NC and two normalizations of the BDM, regarding their robustness when images undergo uniform pixel editing and their behavior when applied to different types of datasets. We found that in terms of uniform pixel editing, the NC is more robust than BDM with the same kind of normalization. The NC is a measure of compression (in this case, using the PAQ compressor) that makes use of the digital object in its entirety to create the shortest possible representation without loss of information. In contrast, BDM divides the digital object into blocks and, based on the complexity of the blocks, estimates the image complexity in its entirety. This means that BDM cannot determine the information shared between the blocks, which causes it to increase, when compared to the NC, with the increase in uniform pixel editing.
In the second test, we compared both measures using different image natures. We found that the results of the NC and NBDM are similar, except for the computed radiography and the cellular automata dataset, which exhibited a more algorithmic behavior. The cellular automata data was created with small programs with simple rules. While the compressor had difficulty compressing this data, BDM could approximate their algorithmic nature and thus assign them a value close to a minimal complexity value. The ability to identify an algorithmic nature incorporated in the data demonstrates the relevance of BDM as a measure.
In the third test, we found that a super sample image transformation causes an underestimation of the amount of information contained in the object by BDM. Again, this is due to BDM analysing the object in blocks, instead of using the object in its entirety. Since the ampliation size was the same as the blocks analysed by BDM, the complexity attributed to each block was approximately zero. Consequently, the overall value attributed to the image complexity decreased dramatically. This aspect demonstrated that BDM cannot handle information contained between each block and can easily underestimate the amount of information present in a digital object.

In the second phase, we applied these measures to estimate the complexity of a dataset of paintings. We calculated the NC and NBDM in this dataset with different quantizations and assessed the results in terms of average complexity per author. Afterward, we combined the NC with the exponent of the roughness of the HDC function in the labeled paintings of the dataset. Finally, we computed the average regional complexity of each author regarding their paintings and built a phylogenetic tree.

We found that paintings with low complexity are abstract, minimalist, and follow simple patterns. Paintings with a slightly higher average complexity possess different regional complexities, specifically, a region with high complexity and detail surrounded by a background of low complexity. With more complexity, this noticeable pattern begins to fade, and the most complex paintings are globally irregular, detailed, and convoluted. Regarding the average complexity values for each artist, we found that NC and NBDM behave similarly, where quantization impacted more the NBDM. We also found that the low side of the complexity spectrum was characterized by Abstract Expressionism, Minimalism, Constructivism movements, with authors such as Mark Rothko, Lucio Fontana, Piet Mondrian, and El Lissitzky. Also, artists from Abstract Expressionism characterized the high complexity side of the spectrum, such as Willem de Kooning, Jackson Pollock, and Jasper Johns, as well as other artists with a more detailed and convoluted style, like Gustav Klimt and Vincent van Gogh. Due to two different currents (Color Field with authors with low average complexity and Action Painting with authors with high complexity), Abstract Expressionism was present at the polar ends of the spectrum. In all cases, Jackson Pollock had average complexity values that were utterly different from other artists, being the average complexity of his paintings close to random. Although he denied being a creator of random paintings, this result and others \cite{Kim-2014a} seem to indicate that Jackson Pollock's dripping paintings are not typical artworks, and this is possibly related to the inclusion of many symbolic layers and dispersion intentions over the canvas by the author.

When evaluating the artists' average NC together with the roughness exponent ($\alpha$) of the HDC function in the labels images of the dataset, we found that styles are well confined into different regions, showing that the combination of these measures gives a robust representation of artistic movements. The NC adds to the level of brightness and relative spatial position evidenced by the roughness exponent, the notion of average information present in each artist's painting, which is consistent within the same style and historical circumstances. We also find that in Abstract Expressionism, the NC is inversely correlated to $\alpha$. Concretely, artists related to Colour Field painting presented a high $\alpha$ and low NC, whereas artists related to Action Painting presented the exact polar results (low $\alpha$ and high NC).

Finally, we divided the image into equal quadrilateral parts and estimated the local complexity of each painting on the dataset and used it to ascertain each artist’s average regional matrix (fingerprint). Complexity can be thought of as a measure of the total number of properties transmitted by an object and detected by an observer. By dividing images into blocks of equal size and evaluating its local complexity, we quantified the local information being transmitted. Furthermore, by averaging the canvas results per artist, we obtain a unique fingerprint that describes how the author exposes information to the observer. Among other things, these fingerprints give specific insights regarding each artist's way of painting, showing where, on average, artists paint with more detail and give more emphasis, while also providing insights into each artist's range of complexity.
Using these matrices, we computed a distance matrix and utilized it to construct a phylogenetic tree.
We discovered that these phylogenetic trees aggregated authors of the same style close to each other, as well as artists' influence relationships, like Francis Bacon and Georges de la Tour, and George Braque and Hieronymus Bosch. 
Furthermore, we observed proximity between artists due to shared methods and techniques which are not correlated with the temporal era or artistic movement. An example of this occurrence is the proximity between Hans Holbein and Vermeer which don't share styles, but both used optics to achieve precise positioning in their compositions.
This evidence shows that artists' fingerprints contain critical information into how the work is perceived, such as composition, unity, balance, movement, rhythm, focus, contrast, pattern, and proportion of the painting and space. 
Finally, we show that these measures improve current methodologies in the classification of artistic paintings and thus extract information which differs from non-handcrafted features. Furthermore, regional complexity provided the largest increase in accuracy on the classification tasks, showing its relevance as a descriptor of images of artistic paintings.

\section*{Conclusions}

In this paper, we introduce novel solutions to the field of computer analysis of artistic paintings and the problem of artist classification and authentication. Specifically, we assessed the viability of unsupervised measures that approximate the quantity of probabilistic and algorithmic information for performing these tasks.
Our direct comparison between NC and BDM allowed us to understand the strengths and weaknesses of both measures. Although BDM has difficulty dealing with uniform pixel edition and full information quantification given the block representability, it serves as a useful tool for measure and indentification of data content having similarity to simple algorithms. On the other hand, the NC is more robust to data alterations (pixel edition and quantization) and is able to measure the quantity of information without underestimation.
Regarding the application of information-based measures in artistic paintings, we studied and developed techniques that can be valuable for art authorship attribution and validation, art style categorization and organization, and art content explanation. Namely, the NC proved to be a robust measure that as a whole gives us some insight regarding the complexity of different styles showing hidden patterns and relationships present in artistic paintings that share the same range in complexity. Furthermore, it could be a stylistic descriptor when coupled with the roughness exponent $\alpha$. On the other hand, fingerprints depict how each author perform typical content distribution on canvas. Thus, they can provide a suitable means of art content explanation, as well as being valuable for art authorship attribution and validation. Moreover, since they provided insights regarding the artists' way of painting, they can be used as a means of relating authors, being therefore useful for depicting artists' stylistic influences, and shared techniques. Additionally, using the distance between the artists' regional complexity, we also find some interesting links between authors regarding the usage of space, technique, composition, rhythm, and proportion. Finally, we demonstrated that the regional complexity and the HDC function of the paintings could serve as useful auxiliary features capable of improving current methodologies in author and style classification of images of artistic paintings.

Regarding future continuations to this study, there are many possible future lines of work that can be considered. For example, in this work we analysed the images of paintings by converting them to monochrome, and it would be interesting to separate the colour channels and to analyse them separately, therefore studying the influence of colour in the paintings in terms of complexity.
Additionally, it would be interesting to explore how to separate different characteristics of the fingerprint, as well as detecting unknown repeated patterns that appear multiple times in a painting by creating and analysing their complexity surfaces \cite{Pratas-2012c}. Lastly, another interesting study would be to replicate the developed work in this article using a competitive compressor that would select the best compressor model for each painting or region.

\section*{Website}

A support website to this site can be accessed at \protect\href{http://panther.web.ua.pt/}{http://panther.web.ua.pt/}. This site showcases among other things, the pipeline of this study, the author's average NC and NBDM variation for different quantization levels, the results of combining the NC with the roughness exponent of HDC function ($\alpha$), a complete catalogue of each author's fingerprints as well as several examples of each author's paintings, and the computed phylogenetic trees with a magnifier to allow a better observation of the results.

\section*{Acknowledgements}
This work was funded by National Funds through the FCT - Foundation for Science and Technology, in the context of the project UID/CEC/00127/2019 and the research grants SFRH/BD/141851/2018 and SFRH/BD/137000/2018 for J.M.S and R.A, respectively. D.P. is funded by national funds through FCT - Fundação para a Ciência e a Tecnologia, I.P., under the Scientific Employment Stimulus - Institutional Call - CI-CTTI-94-ARH/2019.

\section*{Author contributions statement}

J.M.S. and D.P. designed the experiment and wrote the manuscript. J.M.S., D.P., R.A. and S.M. executed the data analysis. All the authors discussed the results and revised the manuscript.

\section*{Additional information}

\subsection*{Competing interests}
 The authors declare no competing interests.


\clearpage
\newpage


\setcounter{table}{0}
\setcounter{figure}{0}
\global\def\thetable{S\arabic{table}}
\global\def\thefigure{S\arabic{figure}}
\renewcommand{\theHtable}{Supplement.\thetable}
\renewcommand{\theHfigure}{Supplement.\thefigure}

\newcommand{\opensupplement}{%
\setcounter{section}{0}%
\renewcommand\thesection{\Alph{section}}%
}

\opensupplement
\section{Supplement}

\subsection{Comparison towards normalized images}
\label{normalization_analysis}
The images provided by the dataset were not normalized. This section evaluates the effects and possible impact of the 8-bit images' normalization on the measures used.
The images were normalized by forcing the brightest pixels to white (255), the darkest pixels to black (0), and spreading the ones in between. We computed each measure's average values per author for the normalized images, and then we measured the average difference and its standard deviation between them and the previously obtained results. Furthermore, we also computed the average percentage difference and standard deviation as 

\begin{equation}
\label{mpd}
\mathrm{Mean~Percentage~Difference} = \sum_{i=0}^{n} \frac{|a_i-b_i|} {   \frac{a_i+b_i}{2}}\times100,
\end{equation}

$a$ and $b$ are the average value of the measure for a given author for the normalized and non-normalized images, respectively,  and $i$ is the author.

Table \ref{tab:normalization} describes the average variation between the measures taken directly from the dataset and those taken after normalization.

\begin{table}[ht!]
\setlength{\tabcolsep}{14pt}
\centering
\caption{Author's Average difference and the percentage difference between normalized and non-normalized images for the NBDM$_1$,NBDM$_2$, NC, and $\alpha$.} 
\label{tab:normalization}
\smallskip
\begin{tabular}{ccc}
\toprule
\textbf{Measure} & \textbf{Average $\pm$ Standard Deviation} & \textbf{Percentage Difference (\%)}\\
\midrule
NBDM$_1$ & 0.031 $\pm$ 0.001 & 3.978\\
NBDM$_2$ & 0.015 $\pm$ 0.001 & 3.977\\
NC & 0.017 $\pm$ 0.002 & 2.531\\
$\alpha$ & 0.001 $\pm$ 0.000 & 0.545\\
\bottomrule
\end{tabular}
\end{table}

The results show that measures have low average variation and percentage differences, being the most affected the NBDM and the least affected the roughness exponent $\alpha$.  Therefore, we can conclude that they are resistant and that the normalization has no significant impact on the measures utilized.

To assess the impact of the image normalization on the Regional Complexity, we performed the Mantel test and computed the average difference between the normalized and non-normalized images' distances. The results are shown in Table \ref{tab:MatelTest}.

The Mantel test measures the correlation between two distance matrices. The results show a high correlation of $0.955$ with a p-value of $0.001$ and a low average difference of distance between authors of $3.2622 \pm 2.820$. Since the only difference between the two distances is that one was computed from normalized images and the other from non-normalized images, we conclude with the results obtained that this measure is also robust to normalization. Consequently, image normalization has a minimal impact and does not significantly influence the measure.

\subsection{Comparison towards other fingerprints}
\label{fingerprint_comparison_section}
\begin{figure*}[ht!]%
\centering
\includegraphics[width=\textwidth]{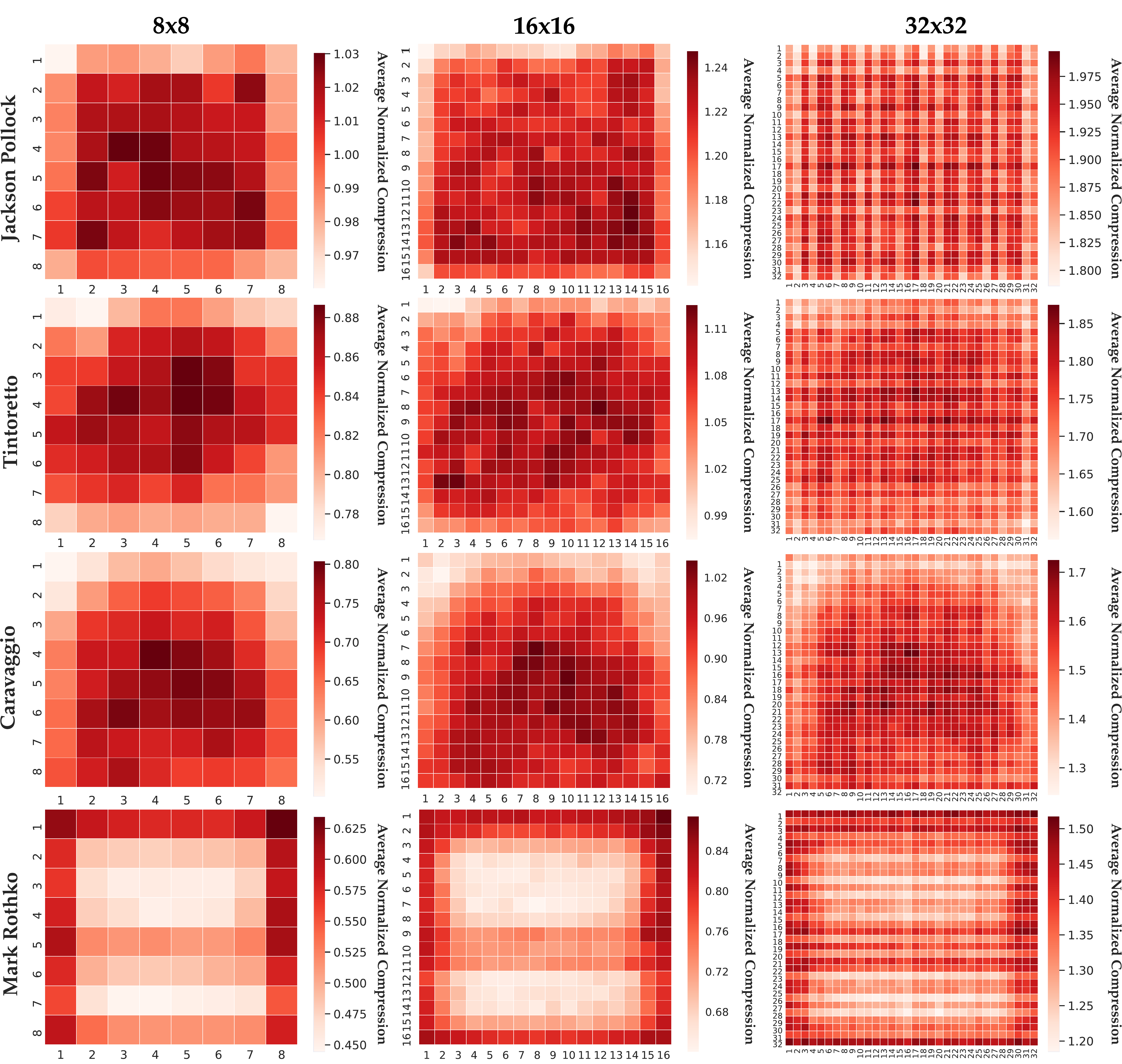}
\caption{Heat maps of the local complexity matrix (fingerprint) of some authors for the different number of blocks the images were divided. This fingerprint shows the author's range of complexity and where they paint with more detail. To see all fingerprints, please visit the website of this article.}
\label{fig:variation_between_blocks}
\end{figure*}

As explained in the Methods section of this article, when creating the fingerprints, we tried to select a patch size that is the minimum for the differences in the compression rate to be significant and capable of being used as a measure between paintings. To this end, we computed the regional complexity of each image for the 8x8, 16x16, and 32x32 blocks of the image. 
It is worth mentioning here that for the regional complexity of 8x8 and 16x16, the division of the image was performed to divide the image into the number of tiles given equally. The result is that the images are divided into tiles which have slightly different sizes. On the other hand, for the 32x32 regional complexities, the images were divided into blocks of the exact same size, except for the last tiles, which were the remain of each image (the remain is the result of images not being an exact multiple of the desired tile size). There are two reasons for this difference in the division method of blocks. Firstly, when we divide the image into a few blocks, each block has a reasonable size and therefore, the variation with the addition of a row or column, does not affect the results of the compression, in contrast with small tiles the addition of a column or a row affects more significantly the compression results of each tile.  Secondly, cropping tiles with the same size in larger blocks would cause a large remainder which would be more inaccurate than to distribute the remainder among the other blocks.
The results of each author's fingerprints (computed for the 8x8, 16x16, and 32x32 blocks) are shown on the auxiliary website of this article. Furthermore, some illustrative results are exemplified in Figure  \ref{fig:variation_between_blocks}.

Qualitatively, as can be seen in Figure \ref{fig:variation_between_blocks} the images of the authors show similar patterns for each author. However, with the increase in patch size, from 16x16 blocks to 32x32 blocks the patterns become less noticeable, as such a good balance between detail and differentiation would be the 16x16 blocks.
Qualitatively, we computed the distance matrix for each author using the images of different patch sizes, and to those distances, we performed the Mantel test and measured the average difference between them. The results are shown in Table \ref{tab:MatelTest}.

\begin{table}[ht]
\setlength{\tabcolsep}{8pt}
\centering

\caption{Mantel Test between distance Matrices and Average difference between them. For the Mantel test, all results had a p-value of 0.001.}
\label{tab:MatelTest}
\smallskip
\begin{tabular}{D{0.30\textwidth}G{0.15\textwidth}G{0.30\textwidth}}
\toprule
\textbf{Comparison methods} & \textbf{Mantel Test} & \textbf{Average $\pm$ Standard Deviation}\\
\midrule
16x16 blocks* & $0.955$ & $\phantom{0}3.262 \pm \phantom{0}2.820$\\
8x8 \textit{vs} 16x16 blocks & $0.951$ & $14.959 \pm 12.372$\\
32x32 \textit{vs} 16x16 blocks & $0.918$ & $74.092 \pm 50.952$\\
\bottomrule\addlinespace[2pt]
\multicolumn{3}{l}{\small* Normalized \textit{vs} non-normalized images.}
\end{tabular}
\end{table}

Since the matrices correspond to distances computed for the same measure but with a different number of blocks, it should be expected for the distance between authors to be similar, thus yielding a high correlation between distance matrices. The results show a high correlation between the distances computed from 8x8 and 32x32 fingerprints towards the 16x16 fingerprints distance. However, there is a higher correlation (0.951) and lower average variation between 8x8 and 16x16 fingerprints distance than between 16x16 and 32x32 fingerprints distance (0.918). This reveals that the increase in the number of blocks from 16x16 to 32x32 decreases consistency between the same author's fingerprints as the correlation between distances decreases significantly. Thus, we can conclude that the 16x16 fingerprints can give a more optimized balance between detail and differentiation since it retains correlation to the 8x8 fingerprints and provides a more detailed map of the author's complexity range.

\subsection{Average \texorpdfstring{NBDM$_2$}{NBDM\_2} per Artist}
\label{nbdm_2_per_Artist}
\begin{figure}[ht!]
\centering
\includegraphics[width=0.5\textwidth]{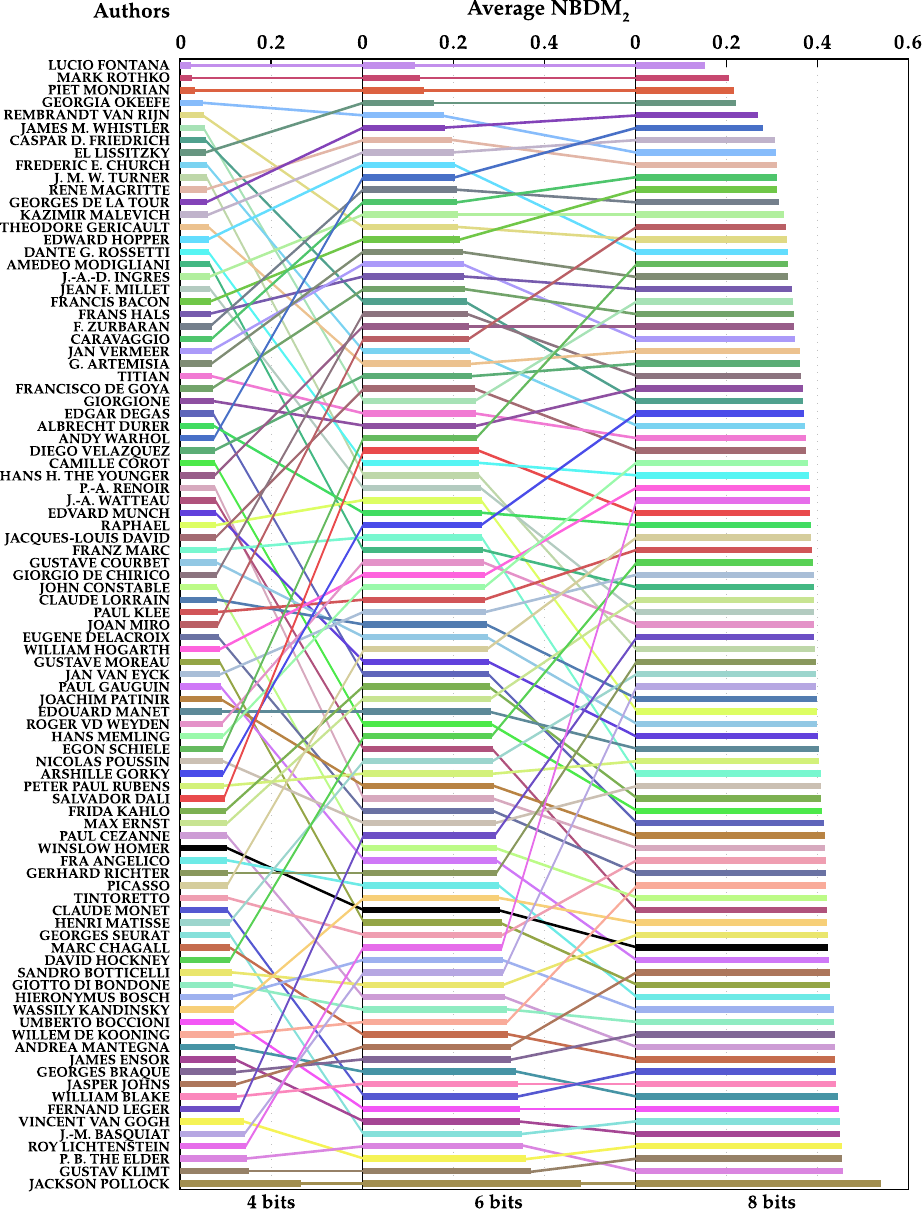}
\caption{Author's average Normalized Block Decomposition Method (ANBDM) using NBDM$_2$  for 4, 6, and 8-bit quantization. The authors are sort given the value of NBDM$_2$. To see this result in more detail, please visit the website associated with the article.}
\label{ABDM2}
\end{figure}

Figure~\ref{ABDM2} shows the average NBDM$_2$ per artist. Each artist has an associated color and its relative positional deviation in different quantizations is illustrated by lines of the same color.

This measure, on average, has a relative positional variation between each author of $13.2 \pm 13.2$, a value slightly lower than in NBDM$_1$, although with a higher average standard deviation. This aspect, combined with the fact that the authors' position varies slightly concerning their position in NBDM$_1$, demonstrates that, overall, normalization has minimal impact on the measure and, thus, it does not influence the results obtained with BDM.

\subsection{Kruskal minimum spanning tree}
\label{Kruskal_section}
\begin{figure*}[ht!]%
\centering
\includegraphics[width=\textwidth]{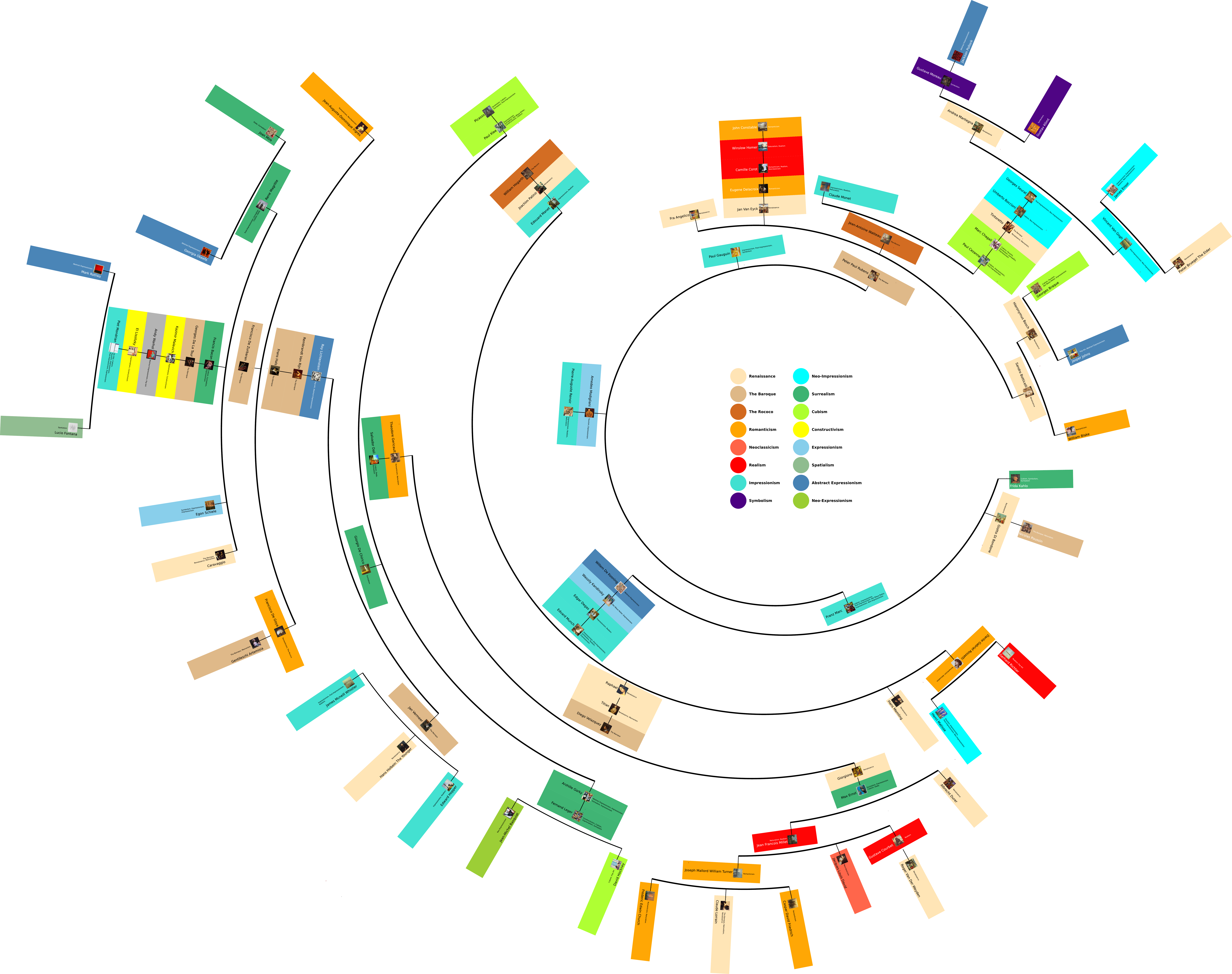}
\caption{Artists' phylogenetic tree computed recurring to Kruskal minimum spanning tree. Each artist has a painting and a color of a style (chosen based on nearest leaves) assigned to him, as well as a description of some styles usually associated with the author. To obtain an improved view of the tree, please see the website associated to the article.}
\label{Kruskal_tree}
\end{figure*}

To verify the congruence of the UPGMA tree, we constructed another tree from the distance tree using the Kruskal minimum spanning tree algorithm \cite{Kruskal-1956}. This algorithm uses the connected graph created by the distance between authors and removes the edges' subset that forms a tree that includes every vertex, where the sum of the weights of all the edges in the tree is minimized.
The resulting tree is shown in Figure \ref{Kruskal_tree} and can also be viewed in more detail on the article's website.

Despite being organized in a different and more sparse manner, the same connections are observed in the UPGMA tree.  The tree Kruskal minimum spanning tree retains the relationships of influence between authors of different artistic movements (Titian and Diego Velazquez; Caravaggio and Francisco de Zurbar\'{a}n; Frida Kahlo and Amedeo Modigliani; Sandro Botticelli and William Blake; Claude Lorrain and Joseph Mallord William Turner; and Peter Paul Rubens and Jean-Antoine Watteau), as well as shows the fingerprint's capacity of grouping some artists from the same artistic movements mutually.
We can conclude from this that qualitatively the fingerprints are useful description tools of the artist's way of painting despite the algorithm used to represent the tree.

\subsection{Replication of Results}

All the results presented in this paper can be fully replicated, under a Linux machine, using the scripts provided at the repository \href{https://github.com/asilab/panther}{https://github.com/asilab/panther}. These include the automatic installation of the tools, download the dataset, assessment, benchmarking, measurement, and visualization of the results. 
\noindent
First, there is the need to give execution access to the scripts using \codeword{chmod +x *.sh}  and perform automatic installation of the tools using \codeword{bash make.sh} and  \codeword{pip3 install -r requirements.txt}

\subsubsection{Information-based Measures Assessment}
\noindent
To download and prepare the dataset, use script \protect\codeword{Dataset.sh}.\newline
To reproduce the compression benchmark, use script \protect\codeword{Benchmark.sh}.\newline
To perform all comparisons between NC, NBDM$_1$ and NBDM$_2$ use \protect\codeword{Compare.sh}.\newline
To replicate the impact of increasing pseudo-random substitutions of pixels for the NC and different types of BDM normalizations (NBDM$_1$ and NBDM$_2$), use script \protect\codeword{Pixel_Edition.sh}.\newline
To test the different values of the NC and NBDM in different datasets use \protect\codeword{Diverse_Images.sh}. If you desire to replicate the cellular automata objects run \protect\codeword{ca.sh}.\newline
To replicate the super-sampling experience and the results of underestimation of BDM, run \protect\codeword{Side_Information_Test.sh}.

\subsubsection{Information-based measures applied to artistic paintings}
\noindent
To perform all the pipeline execute \protect\codeword{Run.sh}.\newline
To quantitize images run \protect\codeword{Quantize.sh} and to trim and binarize use \protect\codeword{Trimm_and_Binarization.sh}.\newline
To compute the average NC, NBDM$_1$, and NBDM$_2$ for each author use scripts \protect\codeword{Average_Complexity.sh}.\newline
To compute the NC with the HDC results use scripts \protect\codeword{NC_HDC.sh}.\newline
To recreate the reports of Regional Complexity, use the following command: \protect\codeword{./Region_Complexity.sh}.\newline
To recreate the reports of author's Fingerprints, use the following command: \protect\codeword{}.\newline
To recreate the authors' fingerprints heat maps run \protect\codeword{Fingerprints.sh}.\newline
To assess the author average variation and percentage difference between normalized and non-normalized measures, use the following command: \protect\codeword{./norm_vs_non_norm.sh }.\newline
To perform the Mantel test and view the average variance between different distance matrices, use the following command: \protect\codeword{./Mantel_test_and_variation.sh }.\newline
To recreate the phylogenetic trees, run \protect\codeword{Tree.sh}. \newline
To make the feature file for author and style classification, run: \protect \codeword{./Create_classification_data.sh}.\newline
To perform author classification, run the jupyter file \protect\codeword{Painting91_author_classification.ipynb}.
To perform style classification, run the jupyter file \protect\codeword{Painting91_style_classification.ipynb}.

\end{document}